# Comprehensive Evaluation of OpenCL-based Convolutional Neural Network Accelerators in Xilinx and Altera FPGAs

R. Tapiador, A. Rios-Navarro, A. Linares-Barranco.
Robotic and Technology of Computers Lab.
University of Seville
Seville, SPAIN
alinares@atc.us.es

Minkyu Kim, Deepak Kadetotad, Jae-sun Seo.
School of Electrical, Computer and Energy Engineering
Arizona State University
Tempe, AZ, USA
jaesun.seo@asu.edu

*Abstract*—Deep learning has significantly advanced the state of the art in artificial intelligence, gaining wide popularity from both industry and academia. Special interest is around Convolutional Neural Networks (CNN), which take inspiration from the hierarchical structure of the visual cortex, to form deep layers of convolutional operations, along with fully connected classifiers. Hardware implementations of these deep CNN architectures are challenged with memory bottlenecks that require many convolution and fully-connected layers demanding large amount of communication for parallel computation. Multi-core CPU based solutions have demonstrated their inadequacy for this problem due to the memory wall and low parallelism. Many-core GPU architectures show superior performance but they consume high power and also have memory constraints due to inconsistencies between cache and main memory. FPGA design solutions are also actively being explored, which allow implementing the memory hierarchy using embedded BlockRAM. This boosts the parallel use of shared memory elements between multiple processing units, avoiding data replicability and inconsistencies. This makes FPGAs potentially powerful solutions for real-time classification of CNNs. Both Altera and Xilinx have adopted OpenCL co-design framework from GPU for FPGA designs as a pseudo-automatic development solution. In this paper, a comprehensive evaluation and comparison of Altera and Xilinx OpenCL frameworks for a 5-layer deep CNN is presented. Hardware resources, temporal performance and the OpenCL architecture for CNNs are discussed. Xilinx demonstrates faster synthesis, better FPGA resource utilization and more compact boards. Altera provides multi-platforms tools, mature design community and better execution times.

*Keywords—Deep Learning; Convolutional Neural Network; Hardware Acceleration; OpenCL; FPGA; Caffe; Xilinx; Altera.*

## I. INTRODUCTION

In recent years, throughout a series of breakthrough algorithms [1-5], convolutional neural networks significantly improved the state-of-the-art in large-scale image recognition tasks. Driven by such success, CNNs have become widespread across a broad range of applications including vision, object detection, speech recognition, autonomous driving, image captioning, etc. Typically CNNs consists of a large number of deep layers, and could involve hundreds of millions of parameters. Using high-end GPGPUs (General Purpose Graphic Processing Units), the networks are trained iteratively using back-propagation algorithm for days or weeks, and then the networks with trained weights can be deployed onto hardware for classification tasks.

There has been a number of prior works [6-12] that built hardware on different platforms for efficient CNN implementation (as accelerators or complete architecture on hardware), such as FPGA [6-9] and ASIC (application-specific integrated circuits) [10-12]. ASIC or custom chip designs show better energy-efficiency, but may not flexibly map various CNN algorithms easily with the rigid circuits. On the other hand, FPGA platforms are much more flexible and could easily map any given CNN algorithm with hardware optimizations. For FPGAs, the designers could perform manual RTL designs [7], but using high-level synthesis tools could prove effective [8-9] in terms of design time and wide design space exploration. The authors in [8] employed HLS tools in Xilinx framework to optimize CNN implementation, while the authors in [9] explored Open Computing Language (OpenCL) based implementation in Altera framework for throughput optimization of CNNs.

Since the high-level synthesis tools are developed differently within different frameworks of Xilinx and Altera, it is difficult to determine which option or FPGA chip would be the best candidate for certain objectives (area, speed, etc.) from the designer's point of view. In this paper, we provide a comprehensive evaluation and comparison of the same CNN using both Xilinx and Altera's OpenCL-based high-level synthesis tool flows. The remainder of the paper is organized as follows. In Section II, the OpenCL programming and models are described. In Section III, Altera's OpenCL design flow and hardware system is discussed, while Xilinx's SDAaccel design flow and hardware platform is presented in Section IV. LeNet-5 ConvNet [20] for MNIST database digits classification scenario is presented in Section V. In Section VI, the hardware results and implementation are compared between the two designs from different vendors in a comprehensive manner. The paper is concluded in Section VII.

## II. OPENCL FOR FPGA

Parallel computing has considerably improved in the last years thanks to the technology scaling favors. From single core CPUs and DSP (Digital Signal Processors), well oriented to single-instruction-multiple-data (SIMD) vectored architectures, computing market changed to multicore chips in the early year 2k when Intel and AMD started to manufacture them. Nevertheless, Rockwell International manufactured the first dual core chip with its version of the 6.502 with two cores in the eighties [14], sharing the chip's pins on alternate clock phases. DSP architecture are oriented to speed up the signal processing using floating point units. Parallelism is obtained thanks to well oriented memory hierarchy and SIMD, very-long-instruction-words (VLIW) and superscalar architectures to maximize the instructions-per-cycle ratio (IPC). In the past decade, parallelism improvement started to be oriented to multi-core architectures for general purpose computers, or many-core to more specific solutions, as GPGPUs. For example, the Tesla K80 accelerator has 4,992 cores with a dual-GPU design that allows up to 2.9 double precision TFLOPS or 8.73 single-precision TFLOPS [21]. Special interest has existed for the FPGAs and SoC framework that includes programmable logic cells oriented to embedded co-design solutions. Recent FPGA technology is highly competitive allowing ASIC emulation with a considerably high resource necessity, such as Stratix 10 (14nm TriGate process, 5.5M Logic Elements, up to 23 TMACS and 10 TFLOPS) [15] or Virtex UltraScale+ (16nm process, 3.7M Logic Cells and up to 21 TMACs) [16].

To implement a given CNN model onto FPGA hardware, we start from the publicly available codes in the Caffe framework [13]. The input image for the CNN model is converted to a text file from Python interface in Caffe and the text file is read from OpenCL host code. Using the Python interface in Caffe, both the input data and weights are extracted and fed to the OpenCL host code, on a batch of input images. The hardware implementation computes till the last inner product layer output and compares it to the expected output from Caffe, to ensure correct functionality. Typically, the CNN models in Caffe are realized using double-precision values for the nodes and the weights. Considering efficient hardware implementation, we first find out how much precision reduction could be achieved while having minimal degradation in the final classification accuracy, and this reduced precision will be used when the OpenCL codes are written.

OpenCL is an open royalty-free standard for general-purpose parallel programming across heterogeneous platforms [17]. Through a programming interface, OpenCL will form a foundation layer of a parallel computing ecosystem of platform-independent tools, middleware and applications. It is very well oriented to graphics rendering pipelines but it is increasing the interest of the FPGA community. OpenCL consists of an application-programming-interface (API) that coordinates parallel computation across heterogeneous parallel processors under the same platform. It has a programming language with a specified computation environment that supports both data and task-based parallel programming models. Therefore, OpenCL is a framework that includes a language, API, libraries and a runtime system to support software development. Four different models define the core ideas behind OpenCL:

*A. Platform Model*

The OpenCL platform consists of a host computer connected to several devices. Each OpenCL device is divided into compute units (CUs). Each CU is divided into processing elements (PE), where computations occur (Fig. 1). The OpenCL application is implemented as both host code and device kernel code. Each part will run in their specific hardware. The host code submits each kernel code as commands from the host to PEs through the memory hierarchy. When PEs of a CU execute the same sequence of instructions, the control flow is called to be *converged*. In this case, single instructions over multiple PEs occurs, which is in fact the same concept of SIMD. If one PE needs a different sequence of instructions, then the control flow is called to be *diverged*. In OpenCL, converged and diverged control flows may occur in the same kernel, providing great flexibility.

*B. Execution Model.*

OpenCL has two units of execution: kernels that execute in one or more platforms, and a host program that executes on a host computer. Kernels execute the computation through work-items (with an associated ID), which are executed in groups (work-groups). The context of what the kernel executes is defined by the host. The host program uses the OpenCL API to create and manage the context. This API has a set of functions that enable the host interaction with devices through a command-queue. There are three main commands: *kernel* (to order the kernel execution), *memory* (data transfer between host and devices) and *synchronization* (synchronize points for order definition between commands). When a kernel-enqueue command submits a kernel for execution, an index space is defined. This index space is called *NDRange* in OpenCL, which corresponds to an *N*-dimensional index space. *N* is 1, 2 or 3. The *NDRange* is decomposed into work-groups forming blocks. It is defined by three integer arrays: *global size* (the extent of the index space in each dimension), an *offset index* (initial value of indices in each dimension), and the *local-size* (size of the work-group in each dimension).

*C. Memory Model*

There are four different memory regions in OpenCL for work-item execution. *Global memory* is where all work-items of all work-groups have to read and write data. These accesses must be cached. *Constant memory* is a region that remains without changes during the kernel execution. The host initializes this memory. *Local memory* is the one used by work-groups locally. It is shared by all work-items. It can be mapped into regions of the global memory. *Private memory* is a memory region that is only visible for a work-item, such that any other work-item cannot access this memory of a particular work-item. Data flow between memory regions is controlled through commands that the host enqueues. The memory consistency is guaranteed in a work-item and between a work-group and all its work-items, but there is no guarantee of memory consistency between different work-groups executing a kernel.

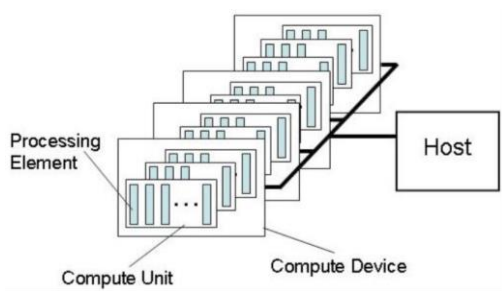

**Fig. 1**: OpenCL platform model.

*D. Parallel Computation Considerations*

In OpenCL, there are mainly two ways to parallelize a kernel: (1) using multiple compute-units (CUs) in parallel (see Fig. 3, left), and (2) vectorising data processing through SIMD kernels with a unique CU (see Fig. 3, right). When multiple CUs are used in parallel, a kernel is replicated and the replicated kernels work simultaneously, increasing throughput and, therefore, consuming global memory bandwidth and hardware resources. On the other hand, SIMD vectorization increases throughput by vectoring kernels, which allows processing multiple work items in a single instruction (SIMD). This alternative is more efficient than using several CUs because it only duplicates the data paths. In this paper, SIMD experiments are presented because the use of replicated CUs generates global memory bottlenecks due to many parallel accesses during the execution.

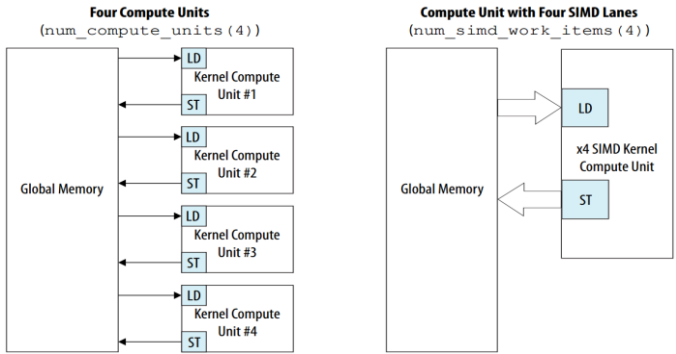

**Fig. 2**: Multiple parallel CU (left) versus single CU with SIMD (right).

*E. Programming Model*

There are two supported programming models: data parallel and task parallel. The data parallel model defines a computation as a sequence of instructions applied to multiple elements of a memory. On the other hand, the task parallel model requires the kernel to be executed in a single work-item of a work-group. In this case, several kernels can be executed in parallel. Synchronization is possible between work-items of a work-group or through two types of enqueued commands: *barrier* (it ensures all previous commands have been executed) or *wait-on-an-event* (the command to be executed waits for a particular event in memory before executing itself).

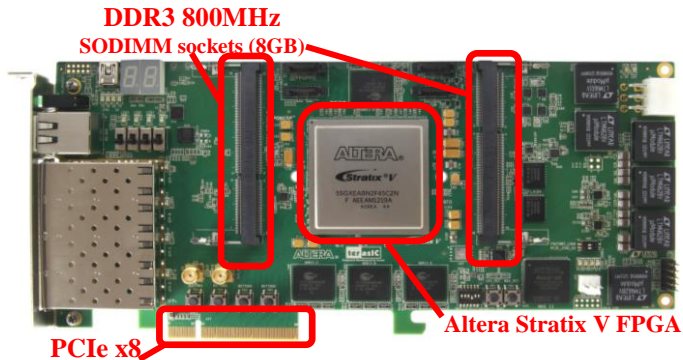

**Fig. 2**: Terasic DE5-Net board.

### III. SCENARIO 1: ALTERA OPENCL

Altera OpenCL (AOCL) is a framework for developing host applications that send kernels to be executed in parallel in FPGA resources. Work-groups, their work-items and memory models are implemented automatically in FPGA resources from an OpenCL description of the kernels and a C++ host application calling API libraries for different functionalities, such as: set buffers, call kernels, synchronize through events and read results. AOCL allows users to abstract the traditional hardware FPGA development flow and instead work with a much faster and higher-level software development flow. Using this design flow, it is possible to emulate OpenCL code in a FPGA, generating synthesis report files as timing or resources summaries.

The design flow consists of two parts: host software application and kernel accelerator hardware on FPGA. The concept is that host sends data to the kernels, where complex calculations are accelerated.

*A. Design Flow*

The design flow for an Altera board using OpenCL consists of several steps. The first step is to describe the functionality of the kernels using C/C++ and then to optimize each kernel applying OpenCL directives to generate a *.cl file. In addition, a host application must be written in C/C++ using the recommended environment.

*B. Hardware Platform*

The implementation of the LeNet-5 CNN for MNIST handwritten digit recognition [20] has been developed on a Terasic DE5-Net (see Fig. 2). This board supports Altera OpenCL and its main characteristics include up to 8 GB DDR3 RAM memory running at 800MHz, 72Mb SRAM running at 550MHz, PCIe-x8 and Altera Stratix V GX-5SGXEA7N-2F45C2 FPGA, which features are shown in Table 1.

**Table 1. Altera Stratix V-GXA7 Specs**

| | |
|---|---|
| Logic Elements (K) | 622 |
| M20K (Blocks / Mbits) | 2560 / 50 |
| 18-bit × 18-bit Multipliers | 512 |
| 27-bit × 27-bit Multipliers | 256 |

### IV. SCENARIO 2: XILINX SDACCEL

The SDAccel [18] development environment is a command line based tool suite for compiling OpenCL programs into a Xilinx FPGA device. The design flow is similar to AOCL in terms of host and kernel descriptions. Directives and API must be replaced when same project is developed for both vendor environments. SDAccel is only available for RedHat Linux OS and it works in a batch mode, where the user invokes the tool with a command file. These commands allow to define the solution name, adding the target device (only one per solution) and host files, creating the kernels and adding the files where they are implemented, creating the Xilinx OpenCL compute unit binary file, and building and packaging the systems. Several CUs per kernel can be implemented. Each CU can have several PEs, which emulates the SIMD architecture. One important advantage over the Altera tool is that SDAccel lets the programmer to test the application before compiling and generating the FPGA binary file. A disadvantage is that SDAccel is less mature than AOCL.

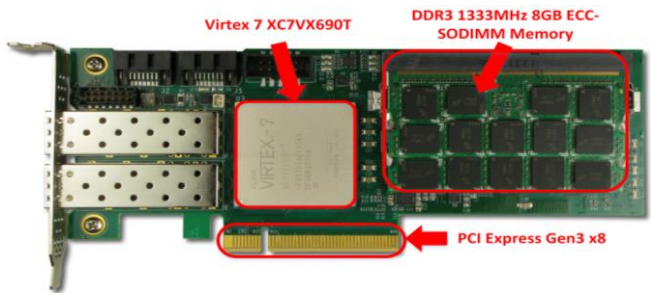

**Fig. 4**: Alpha-Data ADM-PCIE-7V3 board.

*A. CPU and Hardware Emulation*

SDAccel allows emulating in CPU the codesign program before building the system for FPGA. These methods are called CPU and Hardware emulation. The CPU emulation is typically used for functional verification. Each kernel of our solution is compiled into CUs that is executed as a thread on CPU. Hardware emulation is slower since it uses a hardware simulator, but this emulation reproduces the final design on FPGA. The main advantage of using hardware emulation is to avoid the long implementation times (8h on average for this work).

**Table 2: Virtex 7 XC7VX690T features.**

| Physical Characteristics | |
|---|---|
| Logic Cells | 693120 |
| Configurable logic blocks | |
|    Slices | 108300 |
|    Max distributed RAM (Kb) | 10888 |
|    DSP Slices | 3600 |
| Block RAM (blocks/Kb) | 2940/52920 |

*B. Building the System*

This flow builds the system in the real hardware of the target device. When the compilation/implementation is completed, a number of files have been created to run the application, such as the executable, the Xilinx OpenCL binary container (*.xclbin) and the FPGA programming file.

For these experiments, the Alpha Data ADM-PCIE-7V3 [19] board (see Fig. 4) has been used under the CentOS 6.6 operating system. This board is most powerful board that supports SDAccel. The main features include two 8GB ECC-SODIMM memory up to 1333MT/s (faster than Altera DE5 platform), one PCI Express Gen3 x8 and Xilinx Virtex 7 XC7VX690T-2FFG1157C. The features of the FPGA are listed in Table 2. Besides the DSP slices, the specification of the FPGA is similar to that of Altera's Stratix V-GXA7.

## V. LENET-5 AND MNIST SCENARIO

LeNet-5 CNN [20] architecture (shown in Fig. 5) serves as the baseline for many recent CNN-based classification algorithms. It combines three architectural ideas to ensure a certain degree of shift, scale and distortion invariance: local receptive fields, shared weights and spatial sub-sampling. The input layer represents a size-normalized and centered image. In this case, the size corresponds to the size of MNIST database digits (28x28). The first layer is the result of a set of convolutions over the input image. Each pixel in C1 receives inputs from a set of units located in a small neighborhood of the previous layer. This represents the kernel of the convolution. In this exercise, size of the kernel is 5x5. This operation mimics the locally-sensitive, orientation-selective neurons in the cat's visual system, discovered by Hube and Wiesel [21]. These receptive fields in neurons are able to learn and extract elementary visual features, such as edges, end-points, and corners. The combination of these features by subsequent layers are able to detect higher-order features. C1 in this example extracts 20 features from the input image. S2 performs a sub-sampling operation of local averaging, reducing the resolution of the feature maps where distinctive features are encoded. Typically, these convolution and sub-sampling layers are sequentially instantiated for feature map combinations. They are implemented in a bi-pyramid way: at each layer, the number of feature maps is increased as the spatial resolution is decreased. C3 is a convolution layer for 50 smaller feature maps and S4 is the corresponding sub-sampling layer that performs the same operation as that in S2. C3 combines all of the S2 features. The last layer of this CNN is a fully-connected classifier with 500 input neurons and 10 output neurons, which also includes a Rectification Linear Unit (ReLU).

## VI. COMPARISON STUDY

The implementation of this Le-Net5 using the OpenCL framework impose some restrictions. Fig. 5 (bottom) shows the block diagram of the OpenCL solution. It can be seen that the host application, running on a computer, sends input images and kernel weights to the logic through PCIe interface. Data is then stored in the DDR memory in the platform, called "*Global Memory*". This memory is continuously and iteratively accessed by the logic (FPGA) through all the parallel devices physically implemented in hardware. The CNN is structured in 5 kernels (stages), where first kernel implements first layer convolutions and their subsampling operations (conv_pool1); second kernel performs the second layer convolutions (conv2), which is more complex since it has to take results from 20 instances of previous layer, and perform convolutions for 50 instances of this second layer. Then, the third kernel implements the second subsampling operations (pool2). The forth kernel has 500 instances for the classifier unit, whose inputs are the outputs of the previous 50 instances (ip1_relu). Each of these 500 devices send their output to a final layer with 10 instances (one per digit) to categorize the winner digit in the classification (ip2). Each of these devices read the global memory, process the corresponding operation, and then write back the results to the global memory. Consecutive kernels (stages or layers of the CNN) execute in-order, which are controlled by special events included during OpenCL compilations. This architecture needs a high bandwidth DDR memory interface to support all required parallel instances. OpenCL can implement each kernel in a replicated manner as many times in parallel as possible, or it can execute one after the other sequentially if no parallelism can be implemented. As more parallelism is employed, the global memory behavior worsens. The main difference between Altera and Xilinx platforms is the DDR3 on-board memory speed (800MHz for Altera and 1333MHz for Xilinx) as mentioned previously.

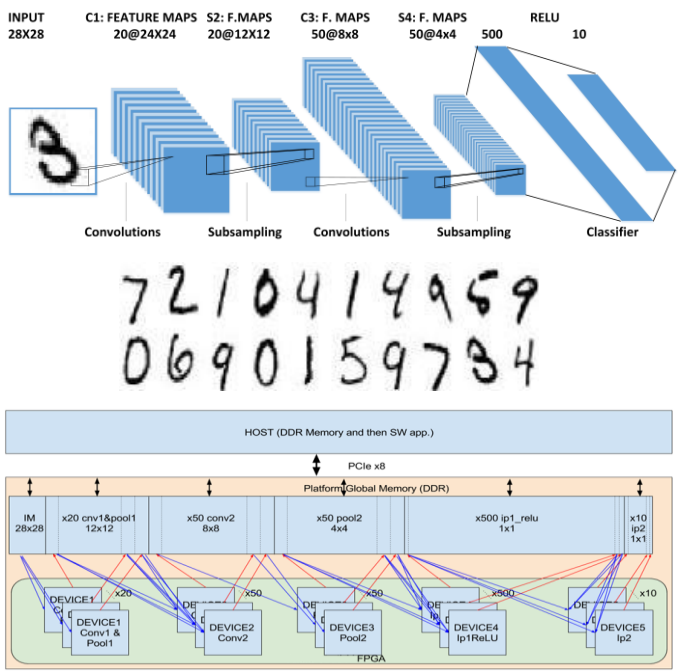

**Fig. 5**: LeNet-5 ConvNet architecture (top) for MNIST digit recognition (middle) and its OpenCL based hardware block diagram (bottom).

OpenCL allows other memory implementations to avoid this shared memory bottleneck, like local pipes that connect two devices directly in the logic. Each of these pipes is implemented through small FIFOs as a point-to-point communication channel between two devices. For CNNs, these pipes do not represent a feasible solution because internal convolution layers, such as C3 in this case, have to read all the S2 outputs and combine them into each of the 50 C3 outputs. This represents 50 pipes at C3 per for each of the 20 S2 units, which is not viable, in terms of resource consumption, for the selected platforms. Therefore, we selected the global memory interface as the possible solution to work for both platforms, and we provide a comprehensive comparison.

Three different tests have been developed for these platforms with the same CNN described in the previous section. The first test consists of comparing each FPGA executing each layer of the CNN without any kind of parallelism. The second test aims to do same measurements when loops are unrolled. For the last test, SIMD directives have been included to vectorise each layer.

Table 3 shows the results of these three experiments for the two vendor platforms. Execution times, logic resources, the number of DSP units and block RAM that are used are shown per kernel. In general, execution time is improved upon employing more parallelism up to a limit. The limit occurs due to the bottleneck that the global memory accesses impose. As expected, the usage of logic gates and DSP units increases when parallelism is increased. Altera tools are able to extract much more parallelism than Xilinx, as it can be seen on logic elements /cells and DSP utilization. There are very small differences between unrolling and SIMD for both platforms for this experiment. Altera tool is able to extract more aggregation for SIMD than Xilinx. In fact, for Xilinx, both unrolling and SIMD have almost same results.

**Table 3: Test results: No parallelism / Unroll / SIMD**

| *Kernel Name* | Execution Time (ms) | Logic Cells /Elem. (K) | DSP slices | BRAM (Kb) |
|---|---|---|---|---|
| *Xilinx Virtex 7 690T* | | | | |
| conv_pool1 | 3.63/1.96/1.96 | 4.9/6.2/5.1 | 11/11/11 | 180/216/216 |
| conv2 | 7.62/4.92/4.92 | 4.8/4.8/4.9 | 11/11/11 | 108/144/144 |
| pool2 | 0.03/0.06/0.06 | 3.0/4.0/3.0 | 4/4/4 | 72/144/144 |
| ip1_relu | 0.55/0.55/0.55 | 4.2/4.2/4.2 | 11/11/11 | 72/72/72 |
| ip2 | 0.35/0.35/0.35 | 4.0/3.0/4.0 | 9/9/9 | 72/72/72 |
| *Altera Stratix V GXA-7* | | | | |
| conv_pool1 | 1.01/1.01/0.98 | 145.7/42.3/73.7 | 8/31/57 | 5225/6205/11200 |
| conv2 | 3.95/3.96/4.27 | 300.5/34.0/34.0 | 8/31/31 | 3207/4882/4900 |
| pool2 | 0.08/0.07/0.13 | 6.9/6.9/6.8 | 2/2/2 | 279/273/279 |
| ip1_relu | 1.01/1.81/2.02 | 5.8/5.8/5.8 | 4/4/4 | 1471/1470/1500 |
| ip2 | 0.15/0.14/0.13 | 5.7/5.7/5.7 | 4/4/4 | 1471/1470/1500 |

**Table 4: Acceleration comparison**

| *Kernel name* | Xilinx vs Altera Acceleration | % Acceleration |
|---|---|---|
| *Conv_pool1* | 3,59/1,94/2 | 259 / 94 / 100 |
| *Conv2* | 1,92/1,24/1,15 | 92 / 24 / 15 |
| *Pool2* | -2,66/1,16/2,16 | - 166 / -16 / -116 |
| *Ip1_relu* | -1,83/3,29/3,67 | -83 / -229 / -267 |
| *Ip2* | 2,33/2,5/2,69 | 133 / 150 / 169 |

In order to demonstrate the different DDR memory bandwidth limits of these two platforms, the same real-time experiment has been performed in both platforms. The experiment consists of connecting a webcam to the host application, which continuously reads in an image frame, normalizing it and resizing to 28x28 pixels using OpenCV libraries. The host sends kernels parameters in the beginning and then it iterates the process of acquiring an image frame, pre-processing it, sending it to the platform and checking the final classification results. The on-board DDR in the Altera platform could not support the memory bandwidth required by this demonstration and the time per frame is continuously increasing. In contrast, Xilinx platform supported this real-time experiment owing to the higher DDR bandwidth. Results show that time increase when parallelism is applied. This is due memory bandwidth when multiple access to global memory are done. Bottlenecks slow down kernel increasing execution time.

In general, logic elements, DSP and BRAM have increased when parallelized directives are applied. However,

the time does not get better due to bottleneck generated by DDR memory bandwidth. Table 4 represents the acceleration between vendors. Execution times for Xilinx are much better than Altera except for *pool2* and *ip1_relu* stages.

## VII. REAL TIME BEHAVIOUR

The previous experiment does not test the behavior against a continuous dataflow of images. In this section a real time experiment has been developed with both platforms (see Fig.6). The experiment uses the OpenCV library in the host computer to capture frames and send them to the FPGA continuously. Altera platform for this particular MNIST experiment, showed a DDR bandwidth bottleneck (800 MHz), which implied a continuously increasing frame processing time, because of the stacked global memory accesses. On the other hand, Xilinx platform didn't reached this DDR bandwidth bottleneck and needed time per frame was constant (DDR memory is 1,3GHz). Nevertheless, Xilinx global cycle time per frame was higher (table 3), but constant.

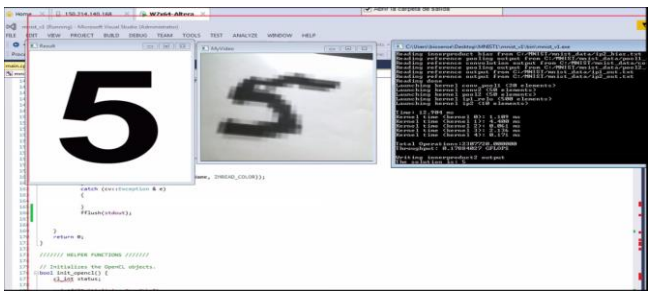

**Fig 6:** OpenCV Application

## VIII. CONCLUSIONS

This work presents a comparison between two OpenCL FPGA-based platforms (Altera and Xilinx) executing a convolutional neural network. Results show that the Altera platform has better execution time for each kernel than the Xilinx platform for all test scenarios. However, the Xilinx platform requires less FPGA resources than the Altera counterpart to execute the same CNN model.

The real-time experiment developed for both platforms has demonstrated that the DDR memory bandwidth is crucial for the global memory communication architecture. Other memory architectures, such as pipes, are implemented internally to the FPGA without requiring any off-chip memory bandwidth, but it was insufficient for CNNs because of their point-to-point restriction. A new memory model that allows having double-buffered memory spread on the FPGA blockRAM will avoid the bottlenecks identified in this work. This will allow having more CUs in parallel to further improve the performance.

ACKNOWLEDGMENT

This work has been partially supported by Samsung Advance Institute of Technology, and by Xilinx and Altera University Programs, through platform donations.

REFERENCES

[1] A. Krizhevsky, I. Sutskever, and G. E. Hinton, "ImageNet Classification with Deep Convolutional Neural Networks," in *Advances in Neural Information Processing Systems*, 2012.

[2] K. Simonyan and K. Zisserman, "Very deep convolutional networks for large-scale image recognition," in *International Conference on Learning Representations (ICLR)*, 2015.

[3] C. Szegedy, *et al.*, "Going Deeper With Convolutions," in *IEEE Conference on Computer Vision and Pattern Recognition (CVPR)*, 2015.

[4] S. Ioffe and C. Szegedy, "Batch Normalization: Accelerating Deep Network Training by Reducing Internal Covariate Shift," in *International Conference on Machine Learning (ICML)*, 2015.

[5] K. He, X. Zhang, S. Ren, and J. Sun, "Deep Residual Learning for Image Recognition," *arXiv:1512.03385*, 2015.

[6] C. Farabet, *et al.*, "Hardware accelerated convolutional neural networks for synthetic vision systems," in *IEEE International Symposium on Circuits and Systems (ISCAS)*, 2010.

[7] V. Gokhale, *et al.*, "A 240 G-ops/s Mobile Coprocessor for Deep Neural Networks," in *IEEE Conference on Computer Vision and Pattern Recognition Workshops*, 2014.

[8] C. Zhang, *et al.*, "Optimizing FPGA-based accelerator design for deep convolutional neural networks," in *ACM International Symposium On Field-Programmable Gate Arrays (FPGA)*, 2015.

[9] N. Suda, V. Chandra, G. Dasika, A. Mohanty, Y. Ma, S. Vrudhula, J. Seo, and Y. Cao, "Throughput-Optimized OpenCL-based FPGA Accelerator for Large-Scale Convolutional Neural Networks," in *ACM International Symposium on Field-Programmable Gate Arrays (FPGA)*, 2016.

[10] Y. Chen, T. Luo, S. Liu, S. Zhang, L. He, J. Wang, L. Li, T. Chen, Z. Xu, N. Sun, and O. Temam, "DaDianNao: A Machine-Learning Supercomputer," in *IEEE/ACM International Symposium on Microarchitecture (MICRO)*, 2014.

[11] Y-H. Chen, T. Krishna, J. Emer, and V. Sze, "Eyeriss: An Energy-Efficient Reconfigurable Accelerator for Deep Convolutional Neural Networks," in *IEEE International Solid-State Circuits Conference (ISSCC)*, 2016.

[12] J. Sim, J-S. Park, M. Kim, D. Bae, Y. Choi, and L-S. Kim, "A 1.42TOPS/W Deep Convolutional Neural Network Recognition Processor for Intelligent IoE Systems," in *IEEE International Solid-State Circuits Conference (ISSCC)*, 2016.

[13] Y. Jia, *et al.*, "Caffe: Convolutional architecture for fast feature embedding," in *ACM International Conference on Multimedia*, 2014.

[14] Rockwell International, "Rockwell R65C00/21 Dual CMOS Microcomputer and R65C29 Dual CMOS Microprocessor," October 1984.

[15] Michael Parker, "Understanding Peak Floating-Point Performance Claims", Technical White Paper WP-012220-1.0, June 2014, Altera Corporation.

[16] UltraScale Architecture and Product Overview, DS890(v2.7), February 2016, Xilinx.

[17] "The OpenCL Specification Version: 2.0," Khronos OpenCL Working Group Editors: Lee Howes and Aaftab Munshi. https://www.khronos.org/

[18] Xilinx SDAccel development environment user guide. http://www.xilinx.com/products/design-tools/software-zone/sdaccel.html

[19] Alpha Data ADM-PCIE-7V3 user manual. http://www.alpha-data.com/pdfs/adm-pcie-7v3%20user%20manual.pdf

[20] Y. LeCun, L. Bottou, Y. Bengio, and P. Haffner, "Gradient-based learning applied to document recognition," *Proceedings of the IEEE*, vol. 86, no. 11, pp. 2278-2324, Nov. 1998.

[21] D. H. Hubel and T. N. Wiesel, "Receptive fields, binocular interaction, and functional architecture in the cat's visual cortex," *Journal of Physiology (London)*, vol. 160, pp. 106-154, 1962.

[22] Tesla K80 GPU Accelerator, Board Specification BD_07317_001_V05, January 2015. http://images.nvidia.com/content/pdf/kepler/Tesla-K80-BoardSpec-07317-001-v05.pdf